\pdfoutput=1 
\documentclass[letterpaper, 10 pt, conference]{ieeeconf}  

\IEEEoverridecommandlockouts                              

\usepackage{graphicx} 

\usepackage{multirow}
\usepackage{algorithmic}
\usepackage{algorithm}
\usepackage{amsmath, amssymb} 
\usepackage{amsxtra}
\usepackage{amsfonts}
\usepackage{url} 
\usepackage{color} 
\usepackage{bm}
\usepackage{placeins}
\usepackage[caption=false]{subfig}
\usepackage{xspace}
\usepackage{mathpazo} 
\usepackage{rotating}
\usepackage{siunitx}
\usepackage{textcomp}

\newtheorem{theorem}{Theorem}[section]

\newtheorem{proposition}[theorem]{Proposition}

\newcommand{\ie}{i.e.,\xspace}

\newcommand{\myvector}[1]{\bm{#1}}
\newcommand{\myvec}[1]{\myvector{#1}}

\newcommand{\R}[1]{\mathbb{R}^{#1}}

\newcommand{\argmax}{\operatornamewithlimits{arg max}}

\newcommand{\algorithmicinput}{\textbf{Input:}}
\newcommand{\algorithmicoutput}{\textbf{Output:}}
\newcommand{\INPUT}{\item[\algorithmicinput]}
\newcommand{\OUTPUT}{\item[\algorithmicoutput]}

\newcommand{\Ly}{Lyapunov }

\newcommand{\excise}[1]{}

\newif\ifremark
\long\def\remark#1{
  \ifremark%
  \begingroup%
  \dimen0=\textwidth
  \advance\dimen0 by -1in%
  \setbox0=\hbox{\parbox[b]{\dimen0}{\protect\em #1}}
  \dimen1=\ht0\advance\dimen1 by 2pt%
  \dimen2=\dp0\advance\dimen2 by 2pt%
  \vskip 0.25pt%
  \hbox to \textwidth{%
    \vrule height\dimen1 width 3pt depth\dimen2%
    \hss\copy0\hss%
    \vrule height\dimen1 width 3pt depth\dimen2%
  }%
  \endgroup%
  \fi}

\newcommand{\TM}[1]{M_{#1}}
\newcommand{\TMM}[2]{M^{#1}_{#2}}
\newcommand{\Tij}{M_{i,j}}
\newcommand{\tm}[2]{m^{#1}_{#2}}

\newcommand{\tijkl}{m^{i,j}_{k,l}}

\newcommand{\X}{X}
\newcommand{\x}{\ensuremath{\myvec{x}}}

\newcommand{\U}{U}
\newcommand{\uv}{\ensuremath{\myvec{u}}}
\newcommand{\ui}[1]{u}

\newcommand{\teta}{\ensuremath{\myvec{\theta}}}
\newcommand{\F}{\ensuremath{\myvec{F}}}

\newcommand{\na}{\ensuremath{d_a}}
\newcommand{\nx}{\ensuremath{d_s}}

\newcommand{\ac}{\uv}

\newcommand{\ftwo}[2]{\|x_{#1} - x_{#2} \|_{0}^2}
\newcommand{\fone}[2]{\text{id}(x_{#1} > x_{#2})}
\newcommand{\dV}{\Delta V(\x, i, j)}

\title{\LARGE \bf
	Resilient Computing with Reinforcement Learning on a Dynamical System: Case Study in Sorting
}

\author{Aleksandra Faust$^{1}$ \and James B. Aimone$^{2}$ \and Conrad D. James$^{2}$ \and Lydia Tapia$^{3}$
	\thanks{$^{1}$Google Brain, Mountain View, CA, USA,{\tt\small faust@google.com}}%
	\thanks{$^{2}$Sandia National Labs, Albuquerque, NM, USA.}%
	\thanks{$^{3}$Department of Computer Science, University of New Mexico, Albuquerque, NM, USA.}%
}
\begin{document}
\maketitle
\thispagestyle{empty}
\pagestyle{empty}

\begin{abstract}
Robots and autonomous agents often complete goal-based tasks with limited resources, relying on imperfect models and sensor measurements. In particular, reinforcement learning (RL) and feedback control can be used to help a robot achieve a goal.  Taking advantage of this body of work, 
this paper formulates general computation as a feedback-control problem, which allows the agent to autonomously overcome some limitations of standard procedural language programming: resilience to errors and early program termination. Our formulation considers computation to be trajectory generation in the program's variable space. The computing then becomes a sequential decision making problem, solved with reinforcement learning (RL), and analyzed with Lyapunov stability theory to assess the agent's resilience and progression to the goal. We do this through a case study on a quintessential computer science problem, array sorting. Evaluations show that our RL sorting agent makes steady progress to an asymptotically stable goal, is resilient to faulty components, and performs less array manipulations than traditional Quicksort and Bubble sort.
\end{abstract}
\section{Introduction}

Modern software controls transportation systems, stock markets, manufacturing plants, and other high-consequence systems. The software often runs in operating environments that are vastly different, and changing, from their original scopes \cite{ackley-cacm-13}. One cause of changes are soft errors, random and temporary errors that affect all aspects of computing, such as memory, registers, and calculations \cite{finocchi-08}. Undetected and unmanaged, their cumulative effect can be severe. For example, particle and electromagnetic radiation corrupt 
computing in space, at high-altitude, and around nuclear reactors, causing silent software failures. Similarly, current microchip design pushes the physical limits, causing memory faults and soft computation errors \cite{caminiti-memory-dp-11}. Traditional radiation hardening and fault-tolerant computing consist of physical system redundancy and material layering. Both increase the complexity and cost.
Resilient algorithms, algorithms that adapt and work around soft errors, are gaining popularity \cite{ackley-cacm-13}. 

Contemporary applications require algorithms to work with limited resources. For instance, an algorithm might be not be allowed to self-terminate. Rather it might be interrupted to yield to a higher priority problem, and must produce a partial answer no worse than the quality of the input data. Thus, an early termination and steady progress toward the solution become requirements.

Yet, traditional computing works under two fundamental assumptions: correctness and program-initiated termination. Theoretical computer science considers an algorithm \textit{totally correct} if, for a correct input, the algorithm terminates and returns a correct calculation \cite{moore-book-11}. The total correctness influences traditional software engineering by developing best practices such as code reviews \cite{hamlet-se-01}, design patterns \cite{bass-softArch-03}, and correctness proofs \cite{hamlet-se-01}, intended to guard against faulty specification and developer-introduced bugs.
The total correctness does not pose any restriction on how computation evolves over time, the result is correct as long as it is given the correct input. That has two consequences. First, the total correctness offers no adaptation and resiliency to errors in the input or the program's dependencies. Second, there are no guarantees on what the algorithm's outcome might be if the program is terminated early.

In contrast, feedback control and sequential decision making in robotics often require error adaptation, and termination when a sufficient accuracy is achieved. Robots routinely rely on measurements that contain errors, yet still aim at providing \textit{resilient} decision making. Next, in many robot control tasks, it is important how a robot accomplishes a task, not only that it does so. For example, among all the trajectories that take the robot from its current position to the goal, we might want to choose the shortest, safest, or most fuel-efficient path. \textit{Stable} decision making finds trajectories that steadily progress towards the goal with respect to some metric. 

This paper applies the \textit{resiliency} and \textit{stability} ideas from controls to computing aiming to overcome correctness and termination limitations. Given the similarities in the operating uncertainties of software and robots, we consider a computation to be a trajectory generation in the program's variable space \cite{moore-book-11}. To that end, we show that a computer program is a discrete time dynamical system over the vector space defined by its variables, and is controlled with vector transformation matrices with an equilibrium in the correct output. Adding a reward, we arrive at Markov Decision Process formulation, which we solve with reinforcement learning to learn a control Lyapunov function. Lyapunov stability theory analyzes the quality of the resulting controlled system, gives theoretical guarantees for steady progression toward the goal, and probability of success in the presence of soft errors. As an example, in order to show how control and decision-making tools can shape algorithm design, we focus on a quintessential computer science problem: array sorting. 
We model the soft errors as an error-prone comparison \cite{ackley-cacm-13,ajtai-impreciseComparison-09}. The resulting agent has the desired resiliency and stability properties.

This paper contributes a description of a sequential state-based computational process as a Markov Decision Process where learning can be applied to produce a stable and resilient solution.  This is achieved through a novel application of reinforcement learning and Lyapunov stability theory to develop a resilient sorting algorithm in environments with a very high probability of soft errors (up to $50\%$).  Specific contributions of this work are 1) a controlled dynamical system formulation for computing in Section \ref{sec:algDynSystem}, 2) 
a RL sorting agent in Section \ref{sec:sorting}, and 3) stability and resiliency analysis of the proposed algorithm (Section \ref{sec:analysis}). The methods presented in this paper are general, and it is expected that they
can be generalized to other iterative, resilient computing algorithms.

\section{Related Work}
\label{sec:relwork}
Sorting has been studied as a quintessential computer science problem in resilient computing. Specifically,  in the context of soft-memory faults due to hardware imperfections \cite{finocchi-08} and ranking applications due to measuring imprecisions \cite{ajtai-impreciseComparison-09}. 
To sort correctly, the first model requires registers that are never corrupted and the known maximum number of memory corruptions \cite{finocchi-08}. 
In contrast, our algorithm requires neither of these. We assume that, at any point, memory corruption is possible with some probability. We model the memory corruption as a $p$-faulty comparison between two elements, which returns an inaccurate comparison with probability $p.$ Our algorithm does not need to know the corruption probability, and the error rate can change over time,  up to $50 \%.$ 
The second error model, ranking, relies on performing multiple rounds on imprecise comparisons. 
Similar to our method, \cite{ajtai-impreciseComparison-09} trades time for accuracy and requires $O (n^2)$ comparisons, while delivering accuracy in probability. RL sort, in addition, prioritizes element selection.

Our solution differs from the previous resilient methods in the applications and tools we use. First, our method makes little to no assumptions about the source and frequency of errors. Faulty comparisons that we choose to work with might be symptomatic of many different causes, such as I/O, transmission, or computation errors. 
Second, our method uses reinforcement learning to learn the near-optimal element selection policy, which is then used to repeatedly choose element to move and its destination.

\textit{Reinforcement programming} \cite{white-RP-12} uses Q-learning to generate a sorting program in absence of errors, using more traditional programming constructs such as counters, \textit{if} statements, \textit{while} loops, etc., as MDP actions. The output is a program ready for execution, tailored for a specific array. Our implementation, in contrast, considers element insertion as the only possible action. Instead of producing an executable program, our agent directly sorts a given array. In other agent-based work, Kinnear uses genetic programming \cite{kinnear-geneticsirt-93}, which produces Bubble sort as its most-fit solution. 

Error measurement of the intermediate results of traditional sorting algorithms reveals a cyclical structure of errors with respect to several metrics \cite{jones-sorting-2014}.
The error oscillations during program execution mean that the results of traditional sorting algorithms, if interrupted before completion, can be worse than the starting state. Our RL sort makes steady progress, and, if interrupted, is guaranteed to return a more sorted array than the one it started with. 

\section{Background}
\label{sec:Preliminiaries}
Trajectory generation over time can be described as a discrete time, controlled, nonlinear dynamical system at time step $n,$
\begin{equation}
D: \;\;\;\;\;\x_{n+1} = \myvec{f}(\x_{n}, \ac_{n}),
\label{eq:dynSystem}
\end{equation}
for a nonlinear function $\myvec{f} : \X \times \U \rightarrow \X$, where the system state $\x \in \X,$ and the input, or action, $\ac \in \U$, influences the system, and changes its current state. Here, we consider fully observable systems, \ie $\myvec{y} = \x.$ %

A deterministic  \textit{Markov decision process} (MDP), a tuple $(\X, \U, D, R)$ with states $\X \subset \R{\nx}$ and action $\U \subset \R{\na}$, that assigns immediate scalar rewards $R:\X \rightarrow\R{}$ to states in $\X$, formulates a task for the system \eqref{eq:dynSystem} \cite{BusBab:10-002}. A solution to a MDP is a control policy $\pi :\X \rightarrow \U$ that maximizes the cumulative discounted reward over an agent's lifetime (state-value function), $V(\x(0)) = \sum_{i=0}^\infty \gamma^i R(\x(i))$, where $0 \leq \gamma \leq 1$ is a discount. System $D$ in \eqref{eq:dynSystem} is a state transition function.

RL solves MDP through interactions with the system and is appropriate when state transition function $D$ or reward $R$ are not explicitly known \cite{BusBab:10-002}. \textit{Approximate value iteration} (AVI) \cite{ernst-avi} finds a near-optimal state-value function, $V:\X \rightarrow \R{}$ 
approximated with a feature map
\begin{equation}
\label{eq:value}
\hat V(\x) = \myvec{\theta}^T\myvec{F}(\x)
\end{equation}
AVI works in two phases, \textit{learning} and \textit{trajectory generation}. The learning phase takes a user-provided feature vector $\myvec F(\x)$ and learns weights $\myvec \theta$ in expectation-maximization (EM) manner. 
After the learning, AVI enters the trajectory generation phase, with an initial state, the feature vector, and the learned parametrization. It creates trajectories using a greedy closed loop control policy with respect to the state-value approximation, 
\begin{equation}
\label{eq:greedyf}
\pi^{\hat V}(\x) = \argmax_{\ac \in A} \hat V(\x'), 
\end{equation}
where state $\x'$ is the result of applying action $\ac$ to state $\x$, $\x' = D(\x,\ac)$. AVI was used in a wide class of problems from control of unmanned aerial vehicles (UAVs) \cite{Ammar_CCA_2010}, to UAVs with a suspended load \cite{faust-ai-13} and electrical power control systems \cite{ernst-avi}, among others.  

\Ly stability theory gives us tools to assess the stability of an equilibrium. 
An equilibrium is \textit{globally asymptotically stable in sense of the Lyapunov} if outcomes of any two initial conditions converge to each other over time \cite{astrom-feedback-08}.
The Lyapunov direct method gives sufficient conditions for stability of the origin \cite{astrom-feedback-08}. The method requires construction of a positive definite scalar function of state $W:\X \rightarrow \R{}$ that monotonically decreases along a trajectory and reaches zero at equilibrium. This function can be loosely interpreted as the system's energy, positive and deceasing over time until it is depleted and the system stops. 
Task completion of a RL-planned motion can be assessed with Lyapunov stability theory, for example, to choose between predetermined control laws in order to guarantee task completion \cite{perkins-lyapunov-2003}, or to construct a state-value function such that it is a control Lyapunov function \cite{faust-ai-13} \cite{faust-acta-13}. We use the latter method here.
\section{Methods}
We first pose general computing as a controlled dynamical system in Section \ref{sec:algDynSystem}. Section \ref{sec:sorting} focuses on a sorting problem, and develops a RL agent, which is analyzed in Section \ref{sec:analysis}.
\label{sec:algDynSystem}

	\subsection{Computing as a Dynamical System}
	\label{sec:methods}
	
	We consider deterministic programs, where a program's outcome is determined by two factors: the initial state of its variables, and the sequence of steps (\textit{algorithm}) that solve the problem. The control-flow constructs, if-then-else, and loops, are controller switches. Instructions are performed at discrete time steps per internal clock tick. The state transitions are determined by executing instructions in the instruction register. 
	A program seen this way is a control policy for a dynamical system determined by the change in the state of variables over time until the computation stops. Computation is a trajectory in the variable space starting with an initial state of variables (initial vector) and terminating at the goal state (goal vector). 
	
	At runtime the program's in-scope variables and the location of the instruction counter uniquely determine the program's state (\textit{configuration}) \cite{moore-book-11}. Regardless of how a program got into a certain configuration, the computation unfolds the same from that point. Thus, a program's state space is the space of all variable values, and satisfies the Markovian property. Without loss of generality, we assume all variables to be real numbers. Thus, a state space for a program with $\nx$ variables is $\X = \R{\nx}$, a $\nx$-dimensional vector. Operations and programming constructs that change the variables, such as \textit{assignment}, \textit{arithmetic operations}, and \textit{changing instructions},  are the action space.  
	Proposition \ref{th:1} shows that in such a setup  a program is a nonlinear dynamical system because the states are vectors and the operations are vector transformations, which can be represented with the transformation matrices. 
	\begin{proposition}
		\label{th:1} 
		A program P with $\nx$ local variables and assignment, summation, and swap operations is a nonlinear discrete time and input system of the form
		$$\x_{n+1} = M \x_n,\;\text{for}\, M \in \U,$$
		where state is a vector $\x_{n} \in \R{\nx},$ and $M \in \U \subset \R{\nx \times \nx}$ is a vector transformation matrix of the state space.
	\end{proposition}
	
	The proof is in Appendix. All programs that manipulate variables are nonlinear control systems per Proposition \ref{th:1}, and the control theory tools can be applied for program analysis. 
	
	Having formulated programs as dynamical systems $(\X,\U,D)$, we only need to provide a reward to formulate MDP \cite{lewis,sutton}. The reward is a scalar feedback on state quality. Typically, the reward is given only when the goal is reached \cite{bartoBook98}, or using a simple metric of a state when available. The next section formulates the reward using sorting as an example.

\subsection{RL Sorting Agent}
\label{sec:sorting}
In this section we develop a stable and resilient RL sorting agent. It learns once, on small arrays, and uses the learned policy to sort an array of arbitrary length. Next, we define MDP and features.

The array sorting state space is the $\nx$-element array itself, $\x = [x_1, \dots, x_{\nx}]^T \in \R{\nx}$. The control space is the discrete set of possible element repositions, $\U = \{(i,j)| i,j=1..\nx\}$. Action $(i,j)$, acts as a list insert. It removes the array's $i^{th}$ element and inserts it into the $j^{th}$ position. Treating arrays as vectors, the actions are permutations of the vector's coordinates, and can be represented with a transformation matrix, $\Tij = [\tijkl ],\,i,j=1\dots{\nx}$. It repositions the $i^{th}$ element to the $j^{th}$ position, $\x' = \Tij \x$, when its elements are defined as \[
\tijkl= \left\{\begin{matrix}
1 & k=l, (k < i\, or \, k > j)\\ 
1 & (k=j, l = i)\, or\, ( i \leq k < j, l=k-1)\\ 
0 & otherwise\\
\end{matrix}\right.
,\; i\le j\; or
\]
\[
\tijkl=
\left\{\begin{matrix}
1 & k=l, (k < j \, or \, k > i)\\ 
1 & (k=j, l = i)\, or \,(j < k \leq i, l=k+1)\\ 
0 & otherwise \\
\end{matrix}\right.
,\ i \ge j.
\]
Matrices $\Tij,$ insert the $i^{th}$ element at the $j^{th}$ position, shift all the elements in between, and do not change elements outside the $[i,j]$ range.

The reward consists of two components: the sum of displaced adjacent elements plus a bonus for reaching a sorted array,
$
R(\x) = \sum_{i=2}^{\nx} (x_{i} - x_{i-1}) \cdot \text{id}(x_{i} - x_{i-1} < 0) + 1000\cdot \text{id}(r_1(\x) == 0), 
$
where $\text{id}(cond)$ equals one when the condition is true, and zero otherwise.

The state-value function approximation, $V$, given in \eqref{eq:value}, is a linear map of a two-dimensional feature vector. We choose features that give an advantage to first sorting areas of the array that are highly unsorted, because our goal is for the agent to perform the most with limited resources. The feature vector is two dimensional. The first feature, $\F_1$, is the number of adjacent out-of-order elements. The second feature ranks arrays with similar elements close together as more sorted than arrays with large variations between adjacent elements. 
\begin{align}
\label{eq:feature}
\F(\x) = & \left[\F_1(\x)\; \F_2(\x)\right]^T \\
= & \left[\sum_{i=2}^{\nx}\fone{i-1}{i} \; \sum_{i=2}^{\nx}\ftwo{i-1}{i}\right]^T
\end{align}
where $\ftwo{i-1}{i} = ((x_{i}-x_{i-1})^2 \text{id}(x_{i}-x_{i-1}<0)).$
To learn $V$, AVI algorithm finds the parametrization $\teta.$ 

Once the feature weights are learned, the RL sorting agent moves to a trajectory generation phase where it sorts arrays without further learning. Instead, at every time step, the algorithm evaluates the current array and chooses an element to move and its new position
\begin{equation}
(i,j) = \argmax_{(k,l) \in [1,..,\nx]^2} {\teta^T \F(\TM{k,l}\x)}.
\label{eq:greedy}
\end{equation}
The chosen action, which maximizes the gain, is applied to the array. The algorithm stops when there are no more displaced elements, \ie the array is sorted. 

\excise{
\begin{algorithm}[h]
	\caption{\small{RL Sort} }
	\label{alg:rlsort}
	\begin{algorithmic}[1]
		\INPUT  $\x_{in}$, an array to sort
		\INPUT  feature vector $\F$
		\INPUT  learned feature vector parametrization $\teta = [\theta_1\; \theta_2]^T$
		\OUTPUT  $\x$, sorted array
		\STATE  $\x \leftarrow \x_{in}$
		\WHILE{$\teta^T \F(\x) <> 0$} \label{ln:while}
		\STATE  $(i,j) = \argmax_{(k,l) \in [1,..,\nx]^2} {\teta^T \F(\TM{k,l}\x)}$ \label{eq:greedy} \COMMENT{find best action to take}
		\STATE $\x \leftarrow \Tij\x$ \COMMENT{place $i^th$ element at the $j^th$ place}
		\ENDWHILE
		\STATE   return $\x$
	\end{algorithmic}
\end{algorithm}  
}
\subsection{Analysis}
\label{sec:analysis}
This section analyzes RL sort. We show that the algorithm is stable, then evaluate its computational complexity and discuss its resiliency.

\subsubsection{Stability Analysis}
\label{sec:stability}
To show RL sort's stability in the absence of errors, we analyze the algorithm's monotonic progression toward the sorted array. The consequence is that the sorted array is an asymptotically stable equilibrium point of the resulting system. 

\begin{proposition}
	\label{th:conv}
	During execution of policy \eqref{eq:greedy} in the absence of errors for an arbitrary array with distinct elements $\x \in \R{n},$ and when both components of the learned weights $\teta$ are negative $(\theta_1 < 0$, $\theta_2<0)$, the following holds:
	\begin{enumerate}
		\item The value function $V(\x)$ increases at every iteration of the algorithm, and
		\item Upon termination of trajectory generation with \eqref{eq:greedy}, the array is sorted.  
	\end{enumerate}
\end{proposition}
The proof, in Appendix, is based on case-by-case analysis of possible scenarios and construction of control Lyapunov function. The proof reveals that RL sort moves elements from the edges into the middle of previously sorted chains, forming increasingly longer and more dense chains. 

The direct consequence of Proposition \ref{th:conv} is that if RL sort gets interrupted, the intermediate result is a more sorted array than the original one. Similarly, the impact of an erroneous comparison sets back the algorithm temporarily, but because of the MDP formulation the algorithm continues with the most current array and without expectations as to how it arrived in that state.

\subsubsection{Computational Complexity of Element Moves}
\label{sec:complex}
RL sort does not modify already sorted arrays. Thus, the lower bound on the number of element moves is $O(1)$, and computational complexity is $O(\nx^2).$
The theoretical upper bound on the number of element moves is $O(\nx^2)$. If the array has $c$ sorted chains, then, in the worst case, there are $\lfloor \frac{\nx}{c}\rfloor$ elements in each, and the elements from the beginnings and endings of all chains are placed in the middle of a single chain, leaving the number of chains and the number of displaced adjacent pairs, $F_1(\x)$, unchanged. After at most  $2*\lfloor \frac{\nx}{c}\rfloor$ element moves we are left with one less chain, and remaining chains have $\lfloor \frac{\nx}{c-1}\rfloor$ elements. Because the maximum number of chains is $c=\lfloor \frac{\nx}{2} \rfloor,$ the conservative estimate of the number of element moves is $\sum_{c=1}^{\lfloor \frac{\nx}{2}\rfloor} 2*\lfloor \frac{\nx}{c}\rfloor = O(\nx^2).$
The empirical results in Section \ref{sec:results} show that this estimate is very conservative, and that in practice the algorithm makes less element moves than Quicksort.  

The computational complexity of policy \eqref{eq:greedy} is $O(\nx^2)$. 
However, action selection with the greedy policy can be improved in several ways. First, a simple way to reduce the computational time to $O(\nx)$, is to use the knowledge gained in Table \ref{tab:options} and restrict the search to only actions that move elements from the edges to the middle of sorted chains. Second, the greedy policy can be parallelized with $O(\nx^2)$ processors, reducing its computational complexity to $O(\log \nx)$ time. With additional $O(\nx)$ storage, the policy evaluation can be done while the elements of the array are being moved. 
Lastly, using specialized hardware acceleration can speed up the action selection by reducing matrix multiplication to linear time, because computation is based on matrix multiplication.
\subsubsection{Resiliency}
\label{sec:resiliency}
When RL sort uses a faulty comparison, the assumptions of Proposition \ref{th:conv} are violated and the stability no longer holds. Thus, this section discusses RL Sort's stability in probability. Specifically, we assess the probability of failing to sort an array and monotonic progression toward the goal. We consider a $p$-faulty comparison component to be an $\text{id}$ function that returns an incorrect answer with probability $0 \leq p \leq 1$, and denote it $\text{id}_p$. Similarly, the feature vector calculated with a p-faulty comparison is denoted $\F_p(\x) = [F^p_1(\x) \; F^p_2(\x)]^T.$ We assume uniform random probability distribution for $\text{id}$ function.

\begin{proposition}
	\label{th:fail}
	Policy \eqref{eq:greedy} that uses a $p$-faulty comparison terminates and fails to sort an array $\x \in \R{\nx}$ with no further processing with probability $P = {\nx \choose k} p^k(1-p)^{(\nx-k)},$ where $k$ is the number of unsorted adjacent elements $k = F_2(\x)$. 
\end{proposition}
The proof is in the Appendix.
The consequence of Proposition \ref{th:fail} is that highly unsorted arrays are unlikely to be recognized as sorted. The probability of terminating by mistake increases as the array becomes more sorted. It also depends on the array size; long arrays are less likely to be taken for sorted.

Next, we discuss the probability that RL sort fails to monotonically progress.
Consider a partition of action set $\U= G \cup N \cup W,$ $G \cap N = G \cap W = N \cap W = \emptyset$.
\begin{align*}
G &= \{(i,j)| \dV > 0\}, \; \|G\| = g,\\
N &= \{(i,j)| \dV = 0\}, \; \|N\| = n,\\
W &= \{(i,j)| \dV < 0\}, \; \|W\| = w,
\end{align*}
where ${\nx}^2 =g+n+w,$ and $\dV = V(\Tij\x) < V(\x)$ is the residual. 
Probability of choosing an action that deceases the value, $\dV < 0$, of the resulting array is a probability of one of the actions from $(i,j) \in W$ ending up having the biggest possible value and being selected. 
Let us denote $p_{V}$ as the probability that the action value changes category ($G,\,N,\,W$) given the comparison's failure rate of $p$. 
The probability of an element from $W$ getting the largest value is, and that value being selected is
\begin{align*}
\text{Pr}((i,j) \in W | V_{p_{V}}(\Tij\x))= &\max_{(k,l) \in A} V_{p_{V}}(\TM{k,l}\x) \\
= &\frac{w}{\nx^2} p_{V}\frac{1}{\nx^2} = \frac{w}{\nx^4}p_{V},
\end{align*}
because the probability that $V_{p_{V}}(\Tij\x)$ is the largest, and therefore selected, is $\nx^{-2}.$

In conclusion, small and almost sorted arrays are more likely to have setbacks while using RL sort because as the array becomes more sorted $w$ becomes larger. We can expect to see no monotonicity violations for highly unsorted arrays, and start seeing more setbacks as the sorting progresses, a trend we see during empirical tests in Section \ref{sec:results}.

When $\nx$ is large, it becomes unlikely that mistakes will have an important impact on the algorithm. Extensive decision-making, a downside from the computational complexity point of view, is an advantage for resiliency. A large number of actions that RL sort examines have a favorable, but not optimal, outcome. Under a favorable outcome, the algorithm selects an action that increases the value, although the increase is not maximal, therefore preserving stability. Traditional sorting algorithms generally perform one array manipulation per decision, impacting their resiliency and stability. Additionally, RL sort is more likely to make less severe mistakes as the probability increases for more sorted arrays.

\section{Results}
\label{sec:results} 
RL sort is compared to Bubble sort and Quicksort, because the two algorithms represent two sorting extremes \cite{ackley-cacm-13}.
The Bubble sort repeatedly scans the list and swaps adjacent out-of-order elements, while Quicksort selects a pivot element and creates three sublists that are smaller, equal, and larger then the pivot. Quicksort then performs a recursive sort on the lesser and greater elements and merges the three lists together.  Quicksort is a single-pass algorithm making large changes in element placement. On the other hand, Bubble sort makes small changes repeatedly until it completes.
The dataset consists of 100 arrays with 10 and 100 uniformly randomly drawn elements. We evaluate RL sort with error-free and $5\%$ faulty comparison. 

\subsection{Learning}
To learn the parametrization $\teta,$ we run the AVI with discrete actions. The samples are drawn uniformly from the space of 6-element arrays with values between zero and one, $\x_s \in (0,1)^6$. The 6-element arrays provide a good balance of sorted and unsorted examples for the learning, determined empirically. We train the agent for 15 iterations. The resulting parametrization, $\teta = [-1.4298 -0.4216]^T$, has all negative elements and meets the conditions in Proposition \ref{th:conv}. 

\begin{table}
	\centering
	\caption{Sorting characteristics demonstrating the impact of random initial distance, the array length, and noise in the comparison routine  averaged over 100 trials. Measures are the number of array element moves.}
	\label{tab:sorting}       
	\begin{tabular} {l|r|rr|rr|rr}
		\hline
		& 
		&\multicolumn{2}{p{1.5cm}|}{0\% Fault}&\multicolumn{4}{p{1.5cm}}{  5\% Fault}\\\cline{3-8}
		Alg. & Len. & 
		\multicolumn{2}{p{1.5cm}|}{\# Moves} &  
		\multicolumn{2}{p{1.5cm}|}{\# Moves} &
		\multicolumn{2}{p{1.5cm}}{Error}\\
		& & $\mu$ & $\sigma$ & $\mu$ & $\sigma$ & $\mu$ & $\sigma$  \\\hline
		RL & 10&10.6 & 2.6 &11.3 & 3.0& 5.3& 23.1 \\ 
		& 100&284.0 & 9.4 &311.3 & 13.0& 0.5& 3.9 \\

		\hline
		\multirow{2}{*}{Bbbl.}
		& 10&23.1 & 4.9&28.0 & 6.7& 3.1& 19.2 \\ 
		& 100&2466.4 &153.2&9836.2 & 992.4& 8.9& 4.8 \\ 
		
		\hline
		\multirow{2}{*}{Quick}
		& 10& 43.8 & 5.3 &42.7 & 4.9& 50.1& 52.0 \\ 
		& 100& 846.9 & 63.4&816.3 & 42.2& 255.8& 74.8 \\ 
		\hline
	\end{tabular}
\end{table}

\subsection{Evaluation}
Table \ref{tab:sorting} summarizes the sorting performance. RL sort finds a solution with the least changes to the array. 
This is because the RL sort does not make the comparisons in the same way traditional sorting algorithms do. Most of its time is spent selecting an action to perform. 
In the presence of a faulty comparison (Table \ref{tab:sorting}), the number of changes to the array that RL sort and Quicksort  perform do not change significantly (less than two standard deviations). The Bubble sort, however, changes the array twice as much. We expect RL sort to seldom make severe mistakes, Quicksort does not reevaluate choices once made, and Bubble sort corrects the mistakes after additional processing. Next, we look into array error. The error is a Euclidean distance, $d(\x^o, \x^s) = \|\x^o - \x^s\|$, between an outcome of sorting with a faulty comparison, $\x_o \in \R{\nx}$, and the reliably sorted array, $\x_s \in \R{\nx}$. No error means that the algorithm returns a sorted array, while high error indicates big discrepancies from the sorted array. Note that this similarity metric would have been an ideal feature vector, but it is impossible to calculate it without knowing the sorted array. With 5\% fault-injection rate, the RL sort's error remains consistent and small across the array sizes, although with a relatively high standard deviation. Bubble sort has a comparable error level but makes an order of magnitude more array changes. The Quicksort completes with an order of magnitude higher error. It is clear that RL sort is resilient to noise and changes the array the least. Table \ref{tab:sorting} shows more comprehensive evaluation results over different datasets, with the similar conclusions.

\begin{figure*}[tb]
	\begin{center}
		\begin{tabular}{ccc}
			\subfloat[Reliable RL sort]{\includegraphics[trim=7mmm 0mm 0mm 0mm, clip,width=0.32\textwidth,height=35mm,keepaspectratio=false]{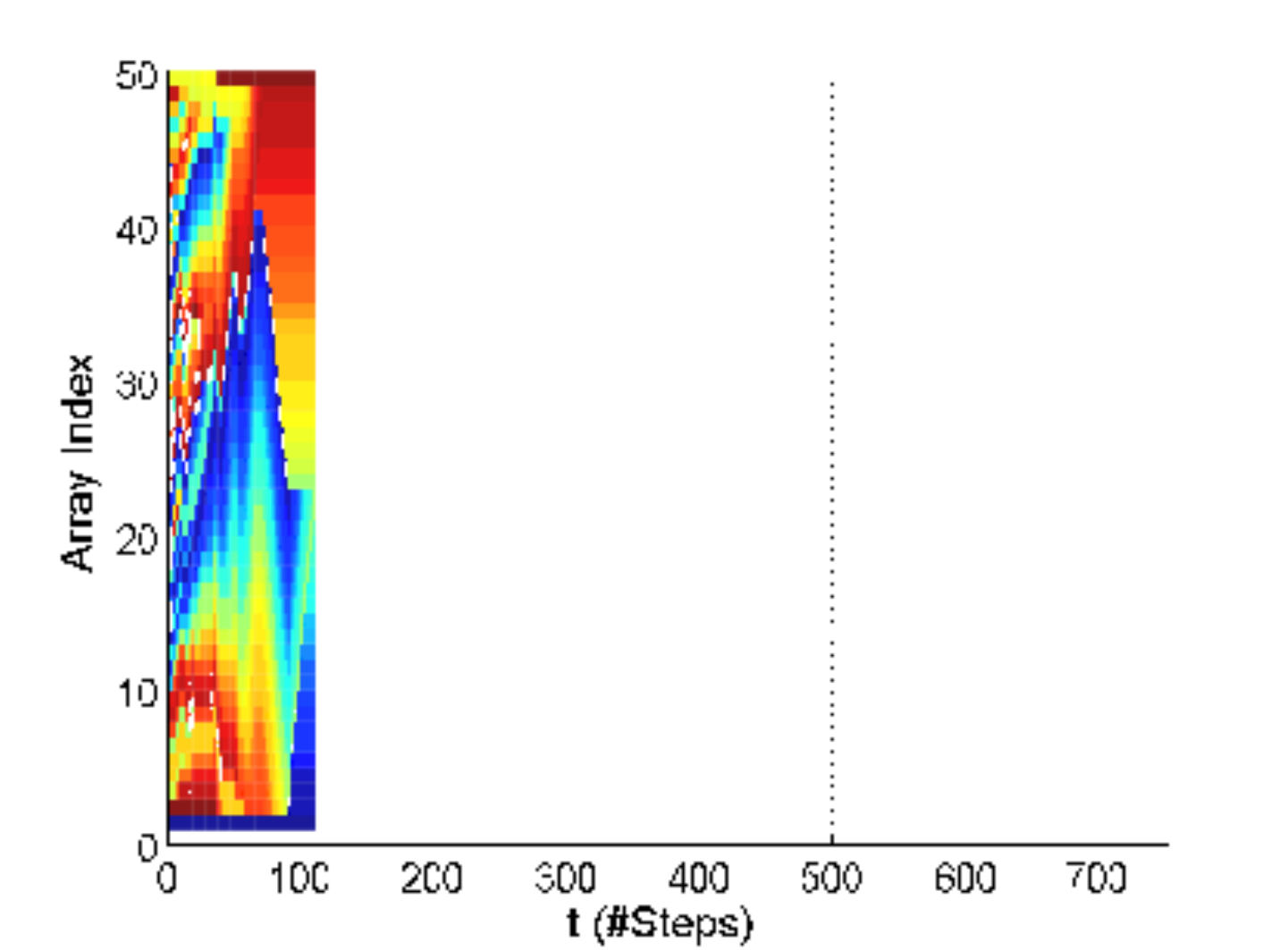}\label{fig:wafr14_sort_RL_r}}&
			\subfloat[Reliable Bubble sort]{\includegraphics[trim=7mmm 0mm 0mm 0mm, clip,width=0.32\textwidth,height=35mm,keepaspectratio=false]{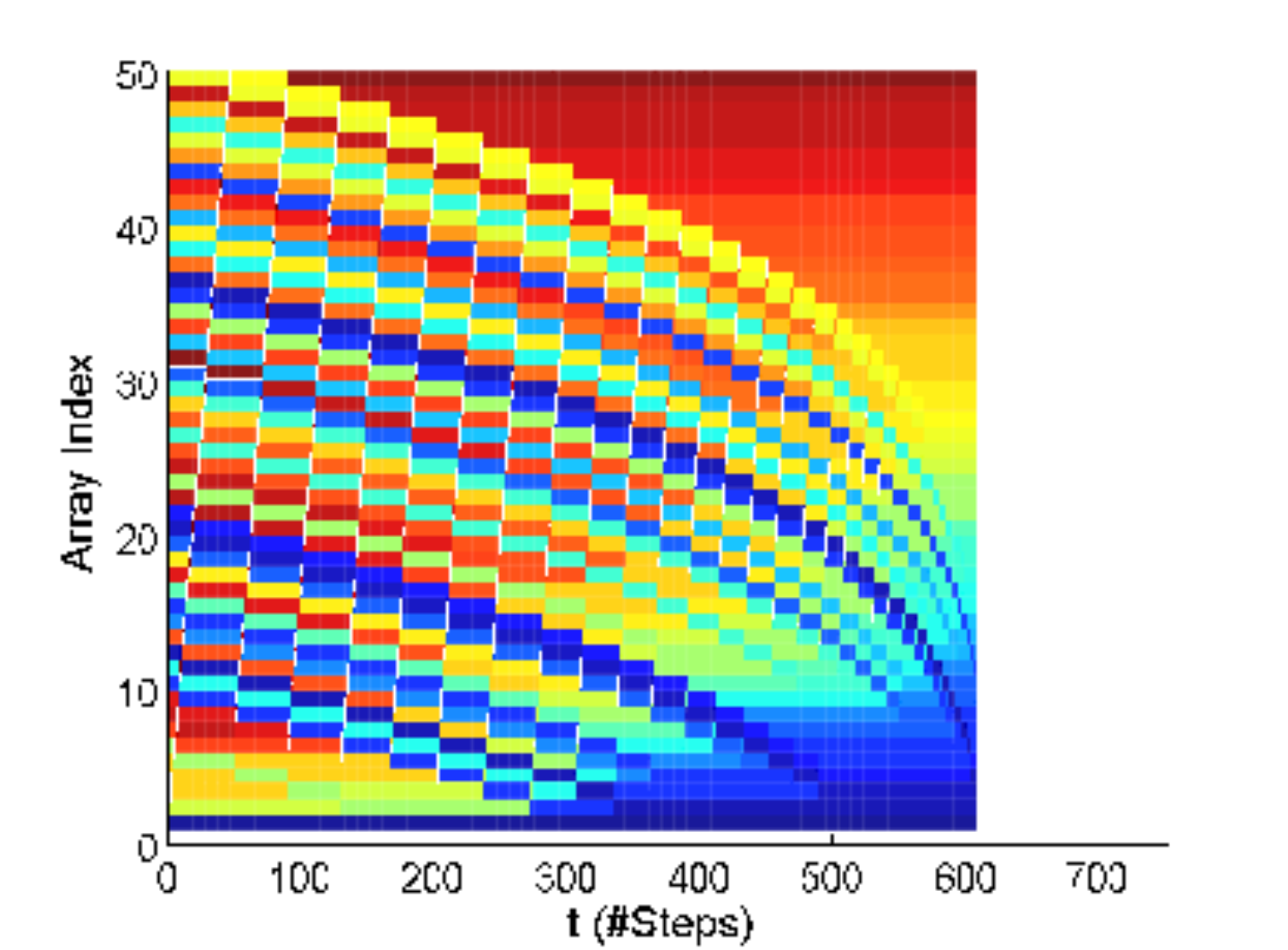}\label{fig:wafr14_sort_bubble_r}} &
			\subfloat[Reliable Quicksort]{\includegraphics[trim=7mmm 0mm 0mm 0mm, clip,width=0.32\textwidth,height=35mm,keepaspectratio=false]{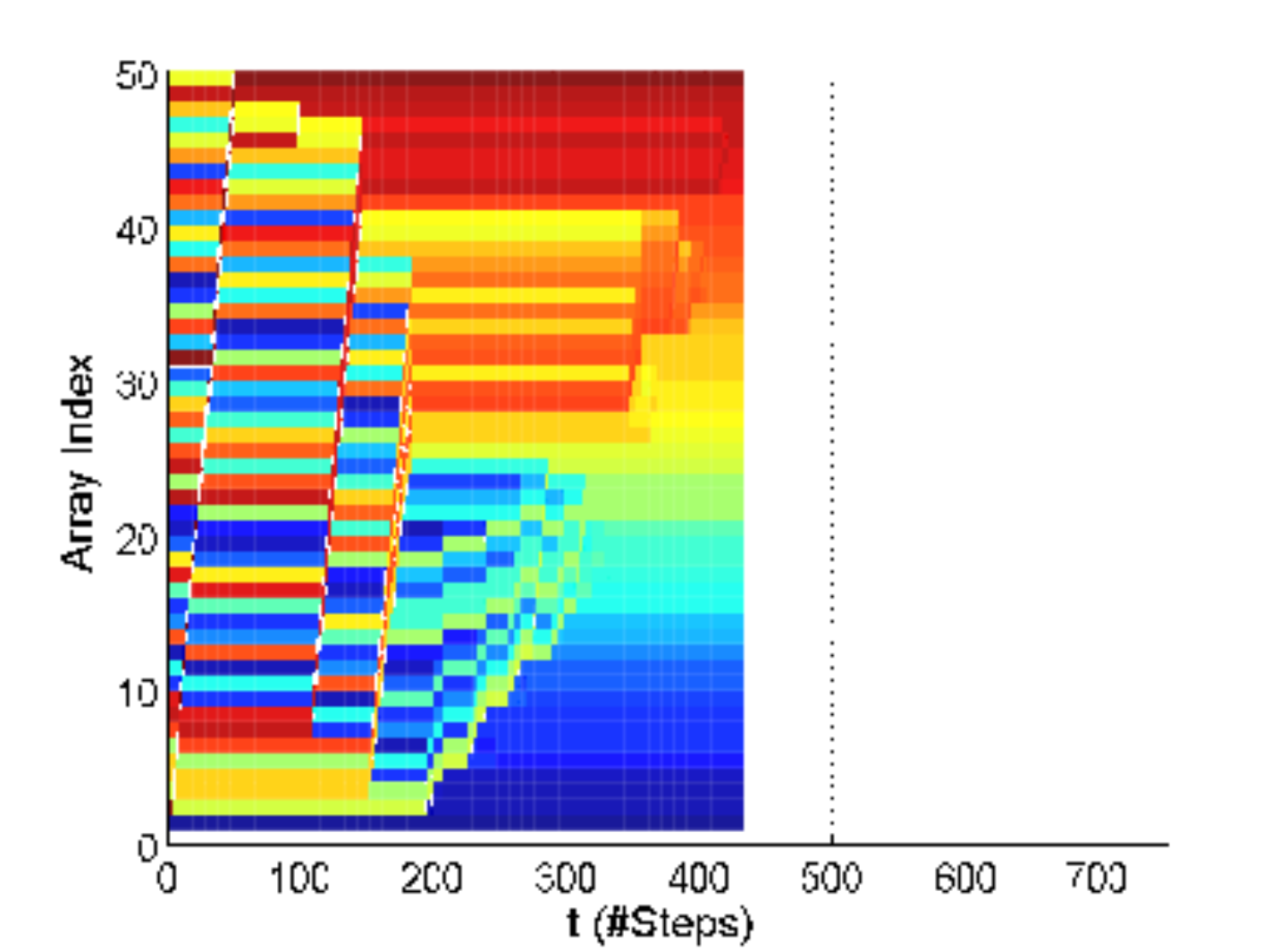}\label{fig:wafr14_sort_quick_r}}\\
			
						\subfloat[Unreliable RL sort]{\includegraphics[width=0.3\textwidth,height=35mm,keepaspectratio=false]{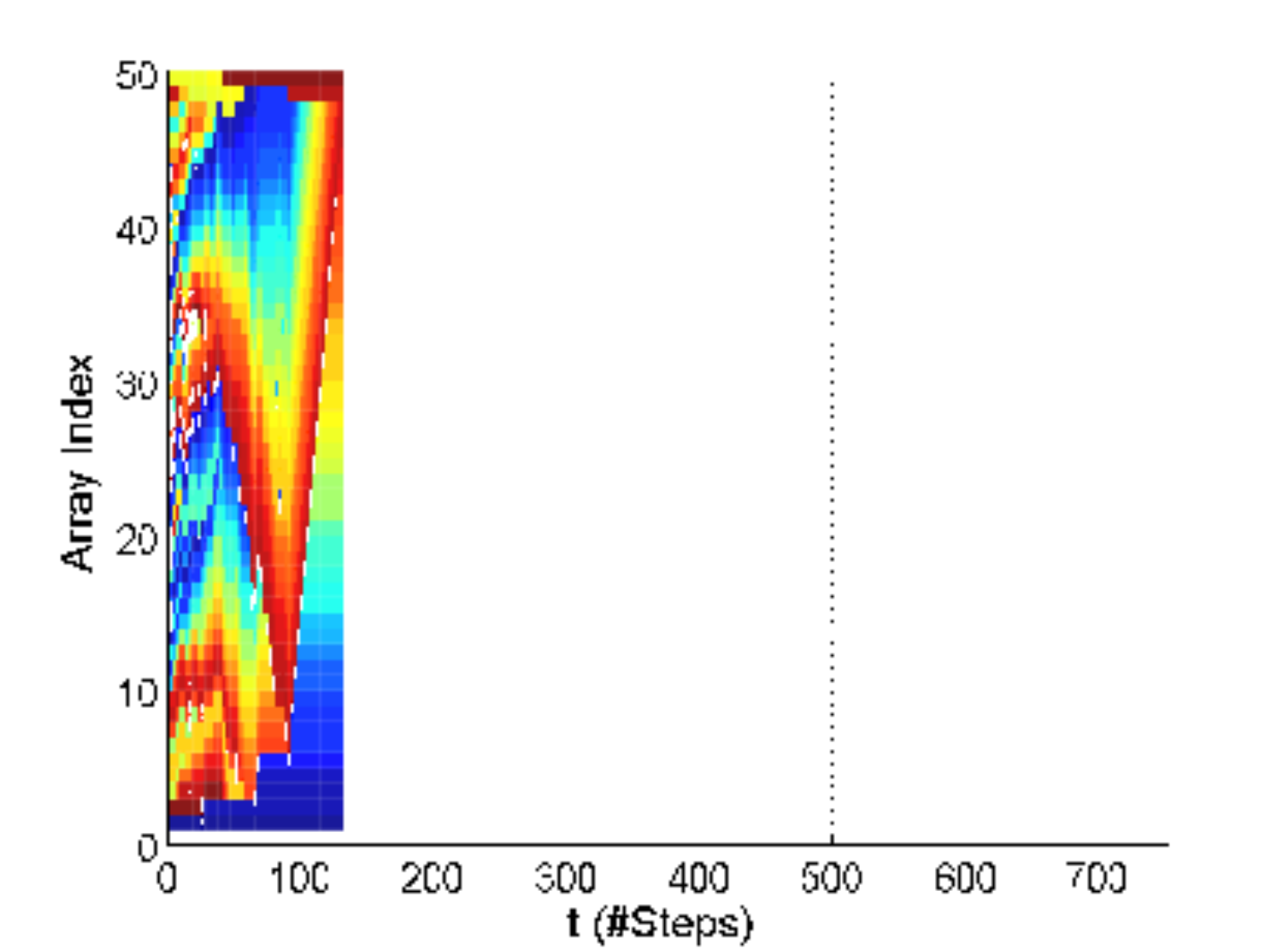}\label{fig:wafr14_sort_RL_n}} &
						\subfloat[Unreliable Bubble sort]{\includegraphics[width=0.3\textwidth,height=35mm,keepaspectratio=false]{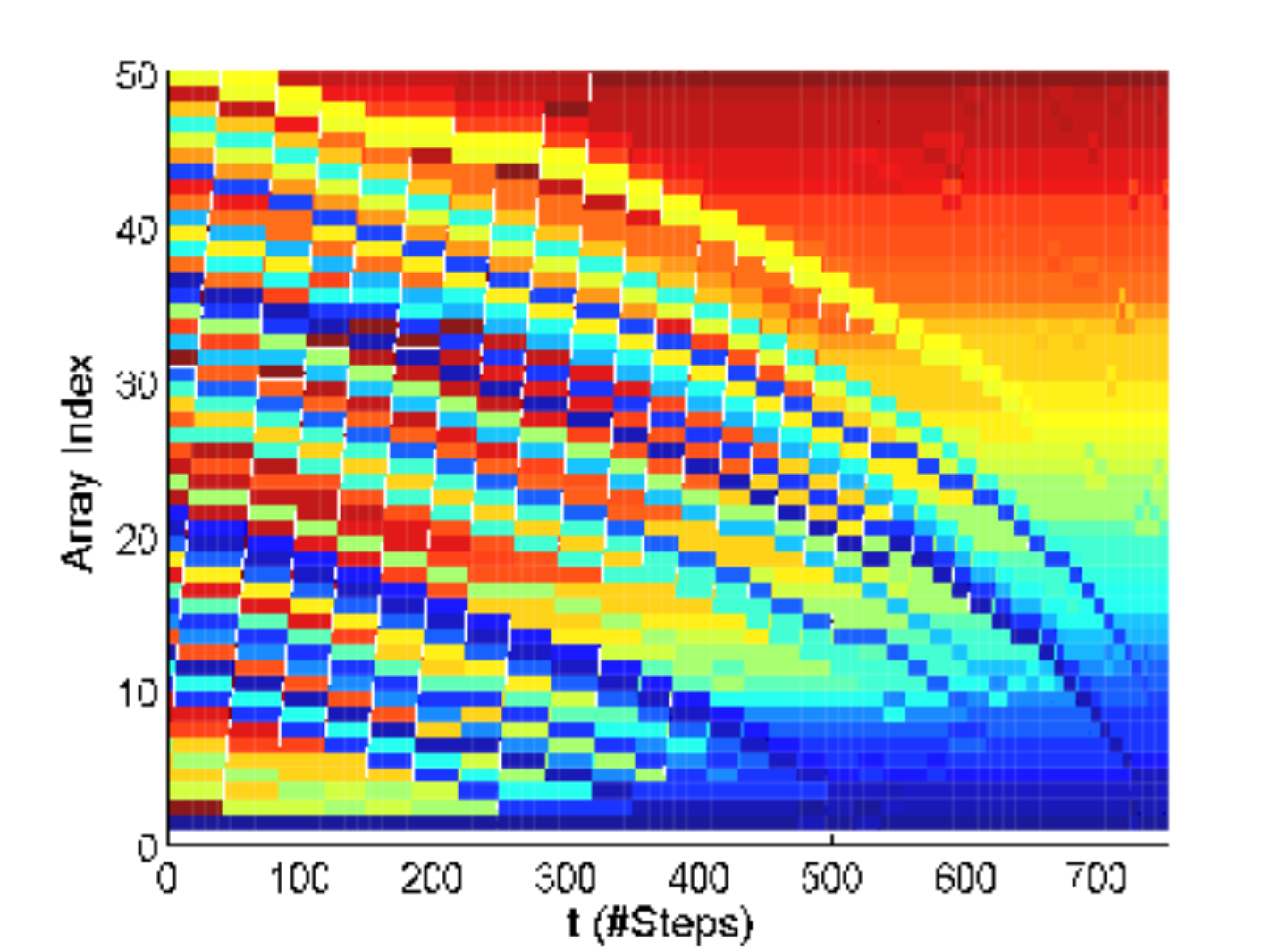}\label{fig:wafr14_sort_bubble_n}} &
						\subfloat[Unreliable Quicksort]{\includegraphics[width=0.3\textwidth,height=35mm,keepaspectratio=false]{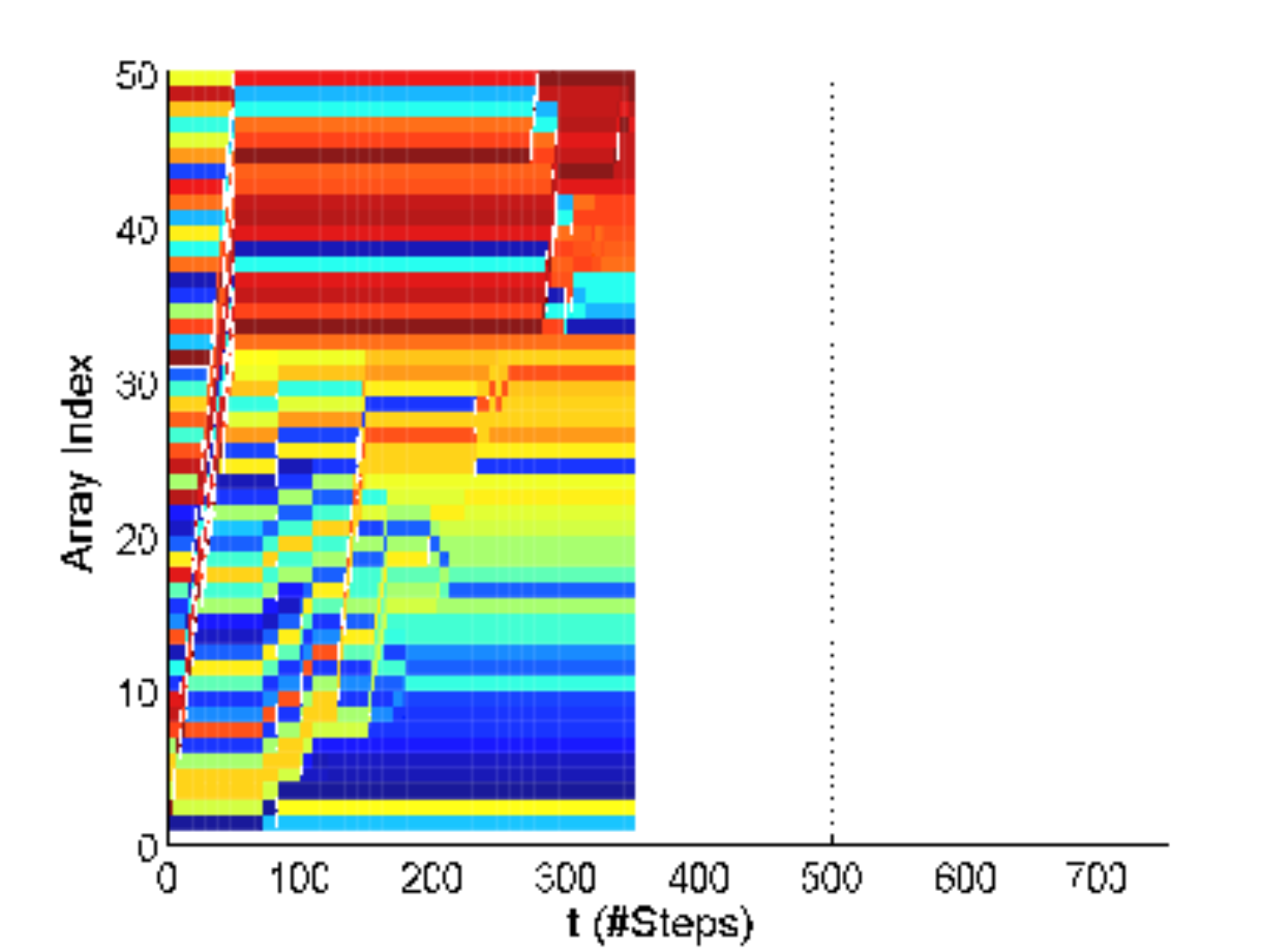}\label{fig:wafr14_sort_quick_n}}\\
		\end{tabular}
		\caption{Sorting progression. A 50-element random array sorted with AVI, Bubble sort and Quicksort with a reliable comparison (a-c) comparison and 5\% unreliable (d-f) comparison components. Time steps are on x-axis, and the array element heatmap is on y-axis. Blue colored are the smallest, and red colored are the largest array elements. Runs end when the array is fully sorted.}
		\label{fig:wafr14_sort_planning}
	\end{center}
\end{figure*}

Fig. \ref{fig:wafr14_sort_planning} visualizes sorting progression of the same array with the three methods, in the absence of errors. Runs end at 111, 433, and 608 steps for RL sort, Quicksort, and Bubble sort, respectively. 
Bubble sort makes small, local changes and Quicksort's movements around the pivot make large-step movements. The RL sort (Fig. \ref{fig:wafr14_sort_RL_r}) takes advantage of the structure in the data: the array is sorted into progressively larger sorted subgroups. This is because the agent reduces the number of adjacent out-of-order elements at each step. Given this, it is no surprise that RL sort needs fewer array manipulations.
Figs. \ref{fig:wafr14_sort_RL_n}-\ref{fig:wafr14_sort_quick_r} depict the same array sorted with unreliable comparison. 
In presence of unreliable comparisons, RL sort takes a different trajectory as the result of the faulty information, it arrives at the goal state, sorted array. The inaccurate information affects Bubble sort locally, because its decisions are local, and it too produces a sorted array. On the other hand, Quicksort, does not revisit sections of the array it previously sorted. Without the redundancy, it fails to sort the array, explaining the high error rates in Table \ref{tab:sorting}.

\begin{figure*}[tb]
	\begin{tabular}{cccc}
	\subfloat[Reliable]{\includegraphics[width=0.23\textwidth,height=3.5cm,keepaspectratio=false]{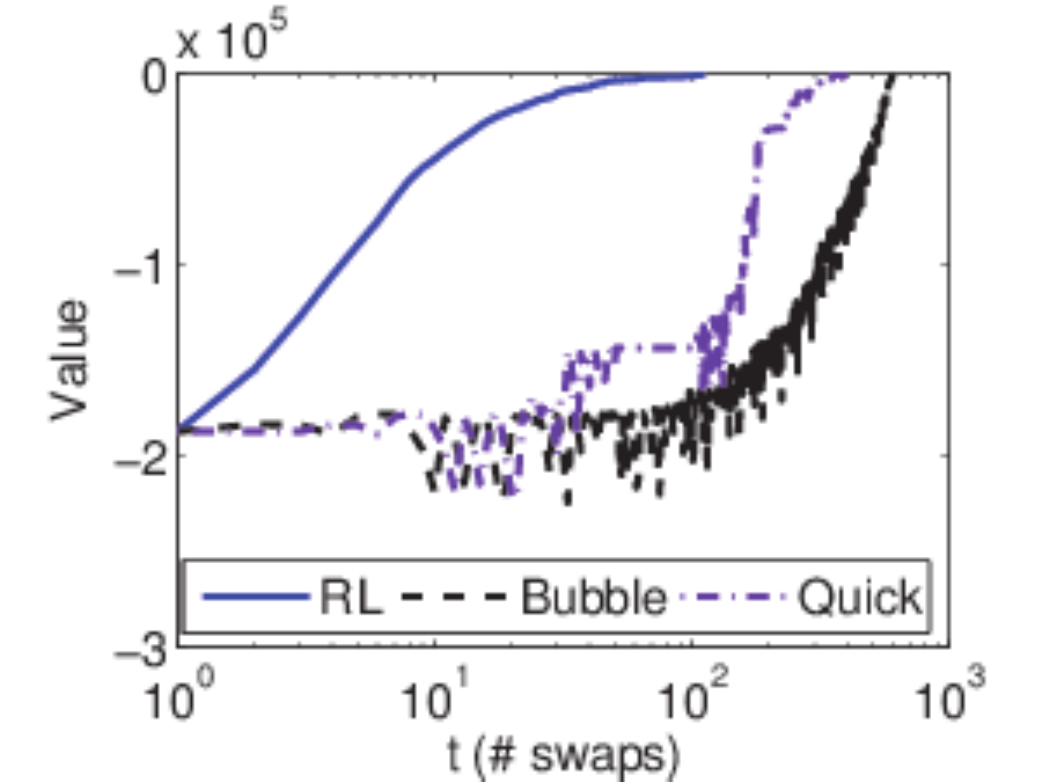}\label{fig:wafr14_sort_value_r}} &
		\subfloat[5\% unreliable]{\includegraphics[width=0.23\textwidth,height=3.5cm,keepaspectratio=false]{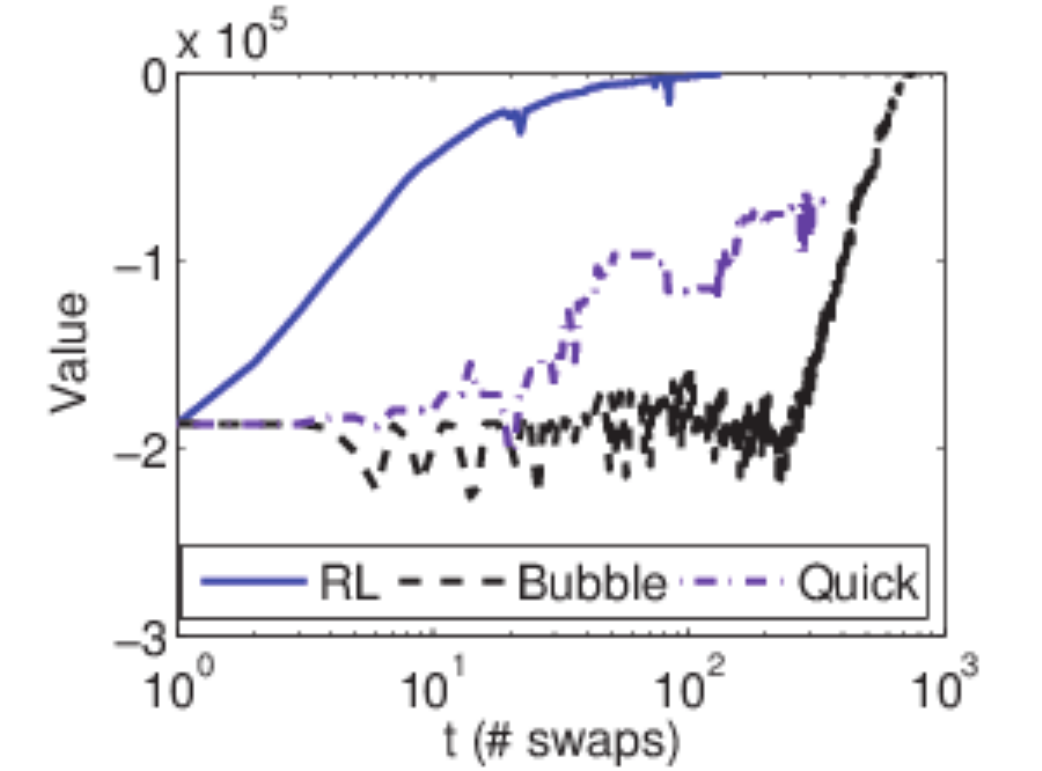}\label{fig:wafr14_sort_value_n}}&
				\subfloat[Success]{\includegraphics[width=0.23\textwidth,height=3.5cm,keepaspectratio=false]{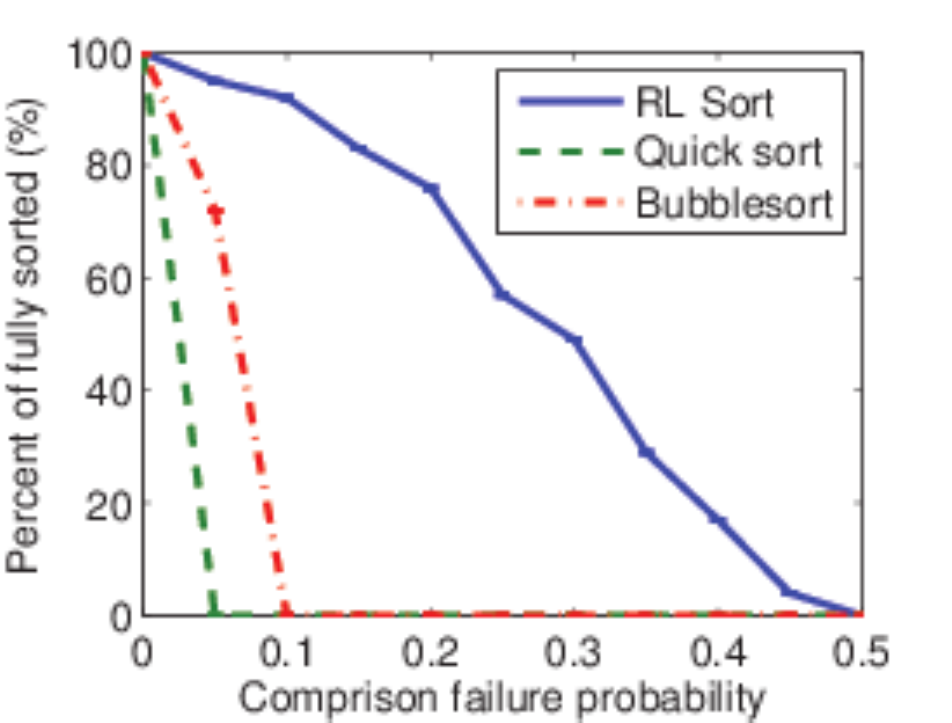}\label{fig:sort15_probSuccess}} &
		\subfloat[Distance from sorted]{\includegraphics[width=0.23\textwidth,height=3.5cm,keepaspectratio=false]{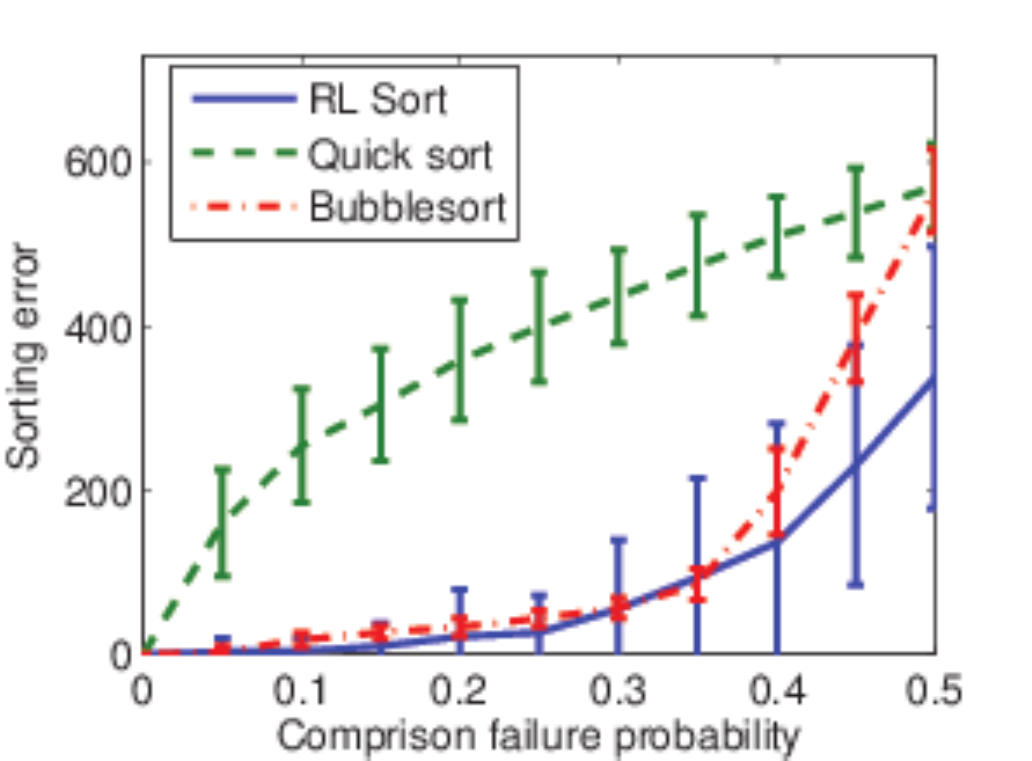}\label{fig:sort_dist}}\\
		
	\end{tabular}
	\caption{(a) and (b) Value progression over time for an array sorted with RL, Bubble sort, and Quicksort with a reliable (a) and 5\%-faulty (b). x-axis is logarithmic. (c) and (d) Percent of successful sorting (a) and average distance from a fully sorted array, and (b) at termination after at varying comparison fault rate.}
	\label{fig:wafr14_sort_value}
\end{figure*}

Visualizing the intermediate array values, $V(\x) = \teta^T\F(\x)$, Figs. \ref{fig:wafr14_sort_value_r} and \ref{fig:wafr14_sort_value_n} offer another view into the algorithms' progression. The RL sort with the reliable comparison monotonically progresses to the sorted array, empirically showing asymptotic stability from Proposition \ref{th:conv}. Bubble sort and Quicksort have setbacks and do not progress monotonically. RL sort with faulty comparison (Fig. \ref{fig:wafr14_sort_value_n}) makes steady progress during the early phases of computation, and experiences temporary setbacks later in the processing, as the analysis in Section \ref{sec:resiliency} predicted. Quicksort fails to reach the same value as RL sort because it stops computation after the first pass. Bubble sort revisits decisions and corrects the faulty decisions, and it eventually reaches the same value as the RL sort.

Lastly, the Figs. \ref{fig:sort15_probSuccess} and  \ref{fig:sort_dist} evaluate resiliency based on the fault rate. Fig. \ref{fig:sort15_probSuccess} shows the percentage of successfully sorted arrays (out of 100) over the failure probability of the comparison routine. For fault rates of 5$\%,$ Quick sort has 0 probability of sorting an array, while Bubblesort fails 100$\%$ of the time when the fault rate is over 10$\%.$ RL Sort has a non-zero probability of sorting an array for fault rates under 50$\%$, confirming our theoretical results. Fig. \ref{fig:sort_dist} measures mean and standard deviation of completion error (Euclidean distance between the terminal state and sorted array). In this graph, lower numbers are better. RL sort has consistently the lowest error. Quicksort's error is a convex, meaning that even small initial errors result in big errors in the terminal state array. Overall, RL sort is more likely to sort an array, and when it fails to sort, the array it produces will be closer to a fully sorted array, than other comparative methods.

\section{Conclusion}
\label{sec:concolution}
This paper presents that \textit{stability} and \textit{resiliency} feedback control and sequential decision making concepts address error and termination limitations of traditional computing. Treating computing as a trajectory generation problem, we apply learning methods to develop an autonomous computing agent, and use Lyapunov stability theory to analyze it.
Specifically, we solve sorting with a \textit{stable} and \textit{resilient} sorting agent.  
We prove the stability in the absence of errors, and discuss the probability of success and to maintaining steady progress in presence of soft errors.

The advantages to resilient computing are that the presented method 1) makes very few assumptions about the source of the error, and 2) does not require manual programming, just a problem formulation. The high computational complexity of the resulting agent improves its resiliency in unreliable conditions.
Its practicality for large-scaled general computing can be improved with use of hardware acceleration and other engineering methods.

An empirical study that compares the RL agent to two traditional sorting algorithms, confirmed the theoretical findings, and showed that the RL sorting agent completes the task with less array manipulations than even the traditional counterparts. In future work, we will develop a tighter upper bound and expected number of array manipulations for RL sort.

\section*{Acknowledgments}

The authors thank Vikas Sindhwani, David Ackley, and Marco Morales. Tapia funded in part by the National Science Foundation under Grant 
Numbers IIS-1528047 and IIS-1553266.
This work was supported by Sandia National Laboratories’ Laboratory Directed Research and Development (LDRD) Program under
the Hardware Acceleration of Adaptive Neural Algorithms Grand Challenge project. Sandia National Laboratories is a multi-mission laboratory managed and operated by National Technology and Engineering Solutions of Sandia, a wholly owned subsidiary of Honeywell International, Inc., for the U. S. Department of Energy’s National Nuclear Security Administration under Contract DE-NA0003525. This paper describes objective technical results and analysis. Any subjective views or opinions that might be expressed in the paper do not necessarily represent the views of the U.S. Department of Energy, the United States Government or National Science Foundation.

\begin{table*}[]
	\centering
	\caption{Value function change ($\Delta V(\Tij\x)$) based on possible action $(i,j)$}
	\label{tab:options}       
	\begin{tabular} {l|c|c|c|l}
		\hline\noalign{\smallskip}
		Placement of the element &\multicolumn{3}{|c|}{Destination placement within a sorted chain ($j$)}&Conclusion\\\cline{2-5} \noalign{\smallskip}
		to be moved ($i$)& Middle & Beginning  & End \\
		within a sorted chain& $x_{j-1} < x_{j}$ & $x_{j-1} > x_{j},$ & $x_{j-1} > x_{j},$ $x_{j-1} > x_i,$\\
		& & $x_i < x_j < x_{j+1}$  &  $x_i < x_{j}$\\
		
		\noalign{\smallskip}\hline\noalign{\smallskip} 
		Middle: $x_{i-1} < x_i < x_{i+1}$  & N/A & N/A  & N/A & Not possible\\  \hline \noalign{\smallskip}
		Beginning & $\theta_1 (\fone{i-1}{i+1} - 1) +$ &  N/A & N/A & $> 0$ because $x_i < x_{i+1}$ \\ 
		 $x_{i-1} > x_{i},$ and $x_i < x_{i+1}$  &$+\theta_2 (\ftwo{i-1}{i+1} - \ftwo{i-1}{i})$&   & &  \\ \hline \noalign{\smallskip}
		End: $x_{i-1} < x_{i},$ and $x_{i} > x_{i+1}$& $\theta_2 (\ftwo{i-1}{i+1} -\ftwo{i}{i+1})$& N/A & N/A & $> 0$ because $x_{i-1} < x_{i}$  \\ \hline
	\end{tabular}
\end{table*}
\appendix
\section{Proof for Proposition \ref{th:1}.}
\label{pr:1}
\textbf{Proof for Proposition \ref{th:1}.}
	\begin{proof}
		The proof is by construction. The variable manipulations are changes in the vector space, transforming one vector to another. Finding a transformation matrix between the vectors proves the proposition. 
		
		\textit{Assignment:} Let $\x=[x_1,...,x_{\nx}]^T \in \R{\nx}$ be a vector representing the current state of variables, and the assignment operation $x_i \leftarrow x_j$ assigns the value of the $j^{th}$ variable to the $i^{th}$ variable. Consider a square $\nx$-by-$\nx$ matrix $\TMM{a}{i,j},$ where its elements $\tijkl, 1\leq k,l \leq \nx$ are defined as follows:
		$$\tijkl = \left\{\begin{matrix}
		1 & (k=i, \; l=j)\, or (k \ne i,\; k=l)\\ 
		0 & otherwise
		\end{matrix}\right.
		.$$ This matrix differs from the identity matrix only in the $i^{th}$ row, where the $i^{th}$ element is zero, and $j^{th}$ is set to one. Then, vector $\x' = \TMM{a}{i,j}\x,$ has $i^{th}$ element equal to $x_j$, and others unchanged. Similarly, $\TMM{c}{i}$, where $\tm{i}{i,i} = c,\, \tm{i}{k,k} = 1,\, k \ne i$ and zero otherwise, assigns constant $c$ to the $i^{th}$ variable.
		
		\textit{Summation:} We show construction of the two-variable summation action matrix. The general case can be shown with induction. Consider action matrix  $\TMM{a}{i,j_{1},j_{2}}$ defined with
		$$\tm{i,j_{1},j_{2}}{k,l} = \left\{\begin{matrix}
		1 & (k=l,\; k\ne i) \, or ( k = i,\; l \in \{j_1,j_2\})\\ 
		0 & otherwise
		\end{matrix}\right.
		$$ 
		for $i,j_1,j_2 \in [1, \dots \nx]$.
		As previously, this matrix differs from the identity matrix only in the $i^{th}$ row, where only elements $j_1$ and $j_2$ are non zero. $\x' = \TMM{a}{i,j_{1},j_{2}} \x $ is a vector where $i^{th}$ element equals sum of $j_1^{th}$  and $j_2^{th}$ elements $\x$. 
		
		\textit{Element swapping:} Lastly, we construct a transformation matrix $\TMM{s}{i,j}$ that swaps $x_i$ and $x_j$. Consider $$\tijkl = \left\{\begin{matrix}
		1 & (k=i, \; l=j)\, or \, (k=j, \; l=i) \, or \, (k=l,\; k\ne i, j)  \\ 
		0 & otherwise
		\end{matrix}\right.
		.$$
		This matrix differs from the identity matrix in that elements $(i,j)$ and $(j,i)$	are ones, while the elements $(i,i)$, and $(j,j)$ are zero.
		
		Finally, when the action space is set of transformation matrices, $\U = \{\TMM{a}{i,j},\TMM{c}{i},\TMM{a}{i,j_{1},j_{2}},\TMM{s}{i,j} |\; i,j=1,\hdots,\nx, c \in \R{}\},$ the variable space manipulation with an action $M \in \U$ is a dynamical system, $\x_{n+1} = M\x_n.$
	\end{proof}

\section{Proof for Theorem  \ref{th:conv}}
\label{ap:proof1}
\textbf{Proof for Theorem  \ref{th:conv}.}

\begin{proof}
	First, we show that $V$ increases at every iteration.
	It is sufficient to show that the residual $\Delta V(\x, i, j) = V(\Tij\x) - V(\x) > 0$ for an arbitrary unsorted array \x.
	Moving the $i^{th}$ element to the $j^{th}$ place changes the array from
	$\x = [x_0, \dots, x_{j-1}, x_{j}, \dots, x_{i-1}, x_{i}, x_{i+1}, \dots x_{\nx+1} ]^T $ to
	\[\Tij\x = [x_0, \dots, x_{j-1}, x_{i}, x_{j}, \dots, x_{i-1}, x_{i+1}, \dots x_{\nx+1} ]^T.\]
	
	To simplify the notation without loss of generality, we append the array with elements $\x_0$ and $\x_{\nx+1}$, such that $\x_0$ is smaller than the $\min \x$, and $\x_{\nx+1} > \max \x$. The two additional elements will not be moved during execution of  policy \eqref{eq:greedy}, because action set \U\ does not contain transform matrices for the two new elements.
	
	Note that the residual $\dV$ depends only on neighborhood of the $i^{th}$ and $j^{th}$ elements 
	\begin{align*}
	\dV &= V(\Tij\x) - V(\x) \\
	&=\theta_1 ( \fone{i}{j-1} + \\
	& + \fone{i}{j} + \fone{i-1}{i+1} -\\
	&- \fone{j-1}{j} -  \fone{i-1}{i}  - \\
	&-\fone{i}{i+1} + \\
	&+\theta_2 ( \ftwo{i}{j-1} +  \ftwo{i}{j} + \\
	&+\ftwo{i-1}{i+1} - \\
	&- \ftwo{j-1}{j} -  \ftwo{i-1}{i}  -  \ftwo{i}{i+1},\\
	\end{align*}
	and when the array \x\ is not sorted $V(\x) < 0,$ since $\theta_1, \theta_2 < 0.$
	
	Consider that action $(i,j)$ is selected at an arbitrary iteration. There are three possibilities for element $x_i$; $x_i$ is in the beginning, end, or middle of a sorted chain. 
	
		\textit{End of chain:} Consider $x_i$ is at the \textit{end} of ascending chain, $x_{i-1} < x_{i},$ and $x_{i} > x_{i+1}$. Then $$\ftwo{i}{i+1} < \ftwo{i-1}{i+1},$$ and 
		$\dV$ in the neighborhood of $x_i$ increases (because $\theta_2 < 0$). Thus, $x_i$ is a candidate to be selected.
		
		\textit{Beginning of chain:} Now, consider that $x_i$ is at the \textit{beginning} of an ascending chain, 
		$$x_{i-1} > x_{i},\,\text{and}\,x_i < x_{i+1}.$$
		Similarly, $$\ftwo{i-1}{i} < \ftwo{i-1}{i+1},$$ $\dV$ in the neighborhood of $x_i$ increases, and  $x_i$ is a candidate to be selected.
		
		\textit{Middle of chain:} Last, consider that $x_i$ is in the \textit{middle} of a sorted, increasing or decreasing chain, i.e. $$x_{i-1} < x_i < x_{i+1}\, \text{or}\, x_{i-1} > x_i > x_{i+1}.$$ Removing $x_i$ does not increase value function in the neighborhood of the $i^{th}$ element.  Assume that \x\ is not sorted, and $x_i$, which is in the middle of a sorted chain, is picked by the greedy policy \eqref{eq:greedy}. If the chain is ascending,
		$$\fone{i-1}{i} = 0,\,\fone{i}{i+1} = 0,\,\fone{i-1}{i+1} = 0.$$
		When the chain is decreasing,
		$$\fone{i-1}{i} = 1,\, \fone{i}{i+1} = 1,$$
		$$\fone{i-1}{i+1} = 1,$$
		$$\ftwo{i-1}{i} + \ftwo{i}{i+1} < \ftwo{i-1}{i+1}.$$
		Thus, removing $x_i$ from a sorted chain does not increase $V(\Tij\x)$ in the neighborhood of $x_i$. Because $\x$ is not sorted, there is at least one element $x_k$ such that $x_{k-1} > x_k$.  Choosing either $x_{k-1}$ or $x_k$ will increase $\dV$ regardless of the direction, therefore $x_i$ that is in the middle of a sorted chain will not be selected for the move. 
		
	Next, we look into the feasibility of the selected element's placement. Like before, the selected element $x_i$, can be placed in the middle, beginning, or end of a sorted chain.
		
		\textit{Middle of chain:} Assume that $x_i$ is moved to \textit{middle} of an ascending chain, $x_{j-1} < x_{j}$. Then $$\fone{j-1}{j} = \fone{j-1}{i} = \fone{i}{j} = 0,$$ and 
		$\dV$ in the neighborhood of $x_j$ does not change with insertion of $x_i$. 
		
		\textit{Beginning of chain: }When $x_j$ is the \textit{beginning} of an ascending chain,
		$$x_{j-1} > x_{j},\,\text{and}\,x_i < x_j < x_{j+1}.$$
		Then, 
		$$\ftwo{j-1}{j} > \ftwo{j-1}{i} + \ftwo{i}{j}.$$
		This action will not be chosen because moving into the middle of a sorted chain results in smaller $\dV$. Note that there are always at least two sorted chains, one at the beginning and one at the end of the array, $x_0 < x_1,$ and $x_{\nx} < x_{\nx+1}$.
		
		\textit{End of chain: }When $x_i$ is at the end of an ascending chain, using a similar argument we conclude that it will not be placed at the end of a sorted array. 
	
	Table \ref{tab:options} summarizes how the state-value function residual changes under different scenarios. A strategy that selects an element from a beginning or end of an ascending chain, and places the element into the middle of another sorted chain, results in a strictly positive change in $V(\Tij\x)$. For an unsorted array such element and its placement can always be found. This proves the first part of Propositions \ref{th:conv}.
	
	Proving the second part of Proposition \ref{th:conv}, that the computation progresses towards a sorted array, is simple using the Lyapunov direct method. Let $$W(\myvec{y}) = - V(\myvec{y} + \x^*),$$ where $\x = \myvec{y} + \x^*,$ and $\x^*$ is sorted $\x$. 
	
	\textit{1) } $W(\myvec{0}) = V(\x^*) = 0$
	
	\textit{2) } $W(\myvec{y})$ is always positive outside of origin, because $V(\x) < 0$ for unsorted arrays.
	
	\textit{3) } 
	\begin{align*}
	\Delta W(\myvec{y}) &= \\
	&= -V(\Tij^T (\myvec{y} + \x^*)) + V(\myvec{y} + \x*) \\
	&= - V(\Tij^T (\x -\x^* + \x^*)) + V(\x -\x^* + \x^*) \\
	&= -\Delta V(\x) < 0,
	\end{align*}
	since we showed earlier that $V(\x)$ increases in subsequent stops, when $\myvec{y} \ne \myvec{0}$.
	
	From the above, $W(\myvec{y})$ is a control Lyapunov function, and $\x^*$ is an asymptotically stable equilibrium. Consequently, the computation with respect to the control policy \eqref{eq:greedy} will ensure that the computation progresses to the sorted array starting at an arbitrary initial array. This proves the total correctness part of the proposition, and concludes the proof.
\end{proof}
\section{Proof for Proposition \ref{th:fail}. }
\label{pr:fail}
\textbf{Proof for Proposition \ref{th:fail}. }
\begin{proof}
Trajectory generation terminates for an unsorted array $\x$, only if faulty comparison causes $\teta^T \F(\x)$ to evaluate as $0$ in \eqref{eq:greedy}. The probability of $\F(\x)$ evaluating as 0 is if all calls to $\fone{i-1}{i}, i=2,\dots,\nx$ return 0. Since there are $k$ adjacent elements that are displaced in \x\, the probability 
	\begin{align*}
	&\text{Pr}(\F_p(\x) = 0) = \\
	&=\prod_{i=2}^{\nx} (\text{Pr}(\text{id}_p(x_{i-1} \leq x_i) | \text{id}(x_{i-1} > x_i))\\ 
	&\cdot \text{Pr}(\text{id}_p(x_{i-1} \leq x_i) | \text{id}(x_{i-1} \leq x_i) ) ) \\
	& = {\nx \choose k} p^k(1-p)^{(\nx-k)},
	\end{align*}
	because there are $k$ unsorted adjacent elements.
\end{proof}

\begin{table*}
	\centering
	\caption{Expended sorting characteristics demonstrating the impact of random initial distance, the array length, and noise in the comparison routine averaged over 100 trials. Measures the number of array element moves. The results cover Selection sort, in addition to Bubble and Quicksort. The Selection sort, selects the minimum and places it at the beginning of the array. The table presents the results for different sorted initial conditions: fully sorted, reverse sorted arrays, sorted with a Gaussian displacement, and uniformly randomly shuffled arrays, over arrays of 5, 10, 50, 100 elements long. The results are consistent across the evaluations.}
	\label{tab:sorting}       
	
	\begin{tabular} {l|l|r|rr|rr|rr}
		\hline
		&\multicolumn{2}{p{2.2cm}|}{}
		&\multicolumn{2}{p{2.6cm}|}{Reliable comparison}&\multicolumn{4}{p{4.5cm}}{  5\% Faulty comparison}\\\cline{2-9}
		Algorithm & Dataset & Array & 
		\multicolumn{2}{p{2.2cm}|}{\# Moves} &  
		\multicolumn{2}{p{2.2cm}|}{\# Moves} &
		\multicolumn{2}{p{2.2cm}}{Error}\\
		& & length& $\mu$ & $\sigma$ & $\mu$ & $\sigma$ & $\mu$ & $\sigma$  \\\hline
		\multirow{9}{*}{RL sort}
				& \multirow{4}{*}{Sorted} 
				 & 5&  1.00 & 0.00&1.00 & 0.00& 2.43& 13.02 \\ 
				&& 10&  1.00 & 0.00 &1.00 & 0.00& 2.12& 15.20 \\ 
				&& 50&  1.00 & 0.00 &1.00 & 0.00& 6.22& 26.57 \\ 
				&& 100&  1.00 & 0.00&1.00 & 0.00& 4.82& 23.27 \\ \cline{2-9}
		& \multirow{4}{*}{Reversed} 
		& 5&  5.00 & 0.00&5.23 & 0.63& 2.30& 20.01 \\ 
		&&  10&  10.00 & 0.00&10.83 & 1.33& 4.99& 20.70 \\ 
				 && 50&  50.00 & 0.00&55.88 & 3.46& 0.36& 3.57 \\ 
		&&100& 100.00 & 0.00&110.31 & 7.00& 3.62& 25.32 \\ \cline{2-9}
		& \multirow{4}{*}{Gaussian}  
		 & 5& 3.54 & 0.91&3.74 & 1.09& 10.12& 71.90 \\ 
		&& 10& 9.83 & 2.34&10.41 & 2.76& 23.52& 111.01 \\ 
		&& 50& 109.45 & 6.21&119.35 & 8.14& 20.58& 101.99 \\ 
		&& 100&273.47 & 9.71&298.01 & 12.12& 25.91& 131.13 \\ \cline{2-9}
		 & \multirow{2}{*}{Random}  
		& 5&3.49 & 1.05 &3.79 & 1.43& 2.35& 20.02 \\ 
		&& 10&10.66 & 2.69 &11.34 & 3.05& 5.31& 23.16 \\ 
				&& 50&112.79 & 7.33 & 123.59 & 9.21& 6.11& 26.02 \\ 
		&& 100&284.02 & 9.42 &311.38 & 13.00& 0.57& 3.97 \\
		
		\noalign{\smallskip}\hline\noalign{\smallskip}
		\multirow{9}{*}{Selection}
		& \multirow{4}{*}{Sorted} 
		& 5& 0.00 & 0.00&0.43 & 0.62& 38.08& 59.78 \\ 
		&& 10& 0.00 & 0.00&1.83 & 1.04& 117.26& 66.88 \\ 
		&& 50& 0.00 & 0.00&31.25 & 3.05& 597.80& 56.18 \\ 
		&& 100& 0.00 & 0.00&79.73 & 3.16& 1009.42& 46.09 \\ 
		\cline{2-9} 
		
		& \multirow{4}{*}{Reversed}  
		& 5& 4.00 & 0.00&3.93 & 0.26& 10.60& 28.41 \\ 
		&& 10& 9.00 & 0.00&8.92 & 0.27& 10.96& 21.41 \\ 
		&& 50& 49.00 & 0.00 &48.99 & 0.10& 9.42& 7.77 \\ 
		&& 100& 99.00 & 0.00&98.94 & 0.24& 8.19& 4.59 \\ 
		\cline{2-9}
		
		& \multirow{4}{*}{Gaussian}  
		& 5& 2.67 & 0.87&2.73 & 0.92& 110.38& 179.76 \\ 
		&& 10& 7.09 & 1.33&7.19 & 1.24& 231.80& 204.31 \\ 
		&& 50& 45.35 & 1.82&45.73 & 1.72& 891.94& 228.69 \\ 
		&& 100& 94.50 & 2.02&95.31 & 1.88& 1802.33& 211.98 \\ 
		\cline{2-9}
		& \multirow{4}{*}{Random}  
		& 5& 2.82 & 0.94&2.90 & 0.92& 23.10& 44.82 \\ 
		&& 10& 7.15 & 1.04&7.34 & 1.00& 53.93& 59.19 \\ 
		&& 50& 45.72 & 1.66 &45.99 & 1.65& 260.58& 65.86 \\ 
		&& 100& 95.04 & 1.82 &95.76 & 1.60& 513.70& 63.03 \\ 
		
		\hline
		\multirow{4}{*}{Bubble}
				& \multirow{4}{*}{Sorted} 
				& 5&0.00 & 0.00&0.52 & 1.29& 0.00& 0.00 \\ 
				&& 10&0.00 & 0.00&1.96 & 3.40& 0.50& 3.53 \\ 
				&& 50&0.00 & 0.00&714.94 & 824.94& 0.66& 2.68 \\ 
				&& 100&0.00 & 0.00&9331.80 & 2369.31& 8.03& 4.39 \\  \cline{2-9}
		& \multirow{4}{*}{Reversed}  
		& 5&10.00 & 0.00&11.02 & 2.22& 3.00& 24.10 \\ 
		&& 10&45.00 & 0.00&50.93 & 4.85& 3.42& 14.57 \\ 
				&& 50&1225.00 & 0.00&2058.29 & 605.55& 1.06& 4.03 \\ 
		&& 100&4950.00 & 0.00&9980.28 & 195.54& 9.97& 4.71 \\ \cline{2-9}
		& \multirow{4}{*}{Gaussian}   
		& 5&5.05 & 2.20 &6.03 & 3.01& 31.54& 143.56 \\ 
		&& 10&21.61 & 5.06&27.27 & 8.29& 8.79& 43.61 \\ 
		&& 50&615.51 & 57.93&1443.22 & 678.96& 2.01& 7.51 \\ 
		&& 100&2461.76 & 144.96&9895.62 & 708.73& 51.65& 45.68 \\ \cline{2-9}
		& \multirow{4}{*}{Random}  
		& 5&5.02 & 2.24&5.85 & 2.96& 6.84& 26.37 \\ 
		&& 10&23.19 & 4.95&28.08 & 6.73& 3.13& 19.21 \\ 
				&& 50&609.82 & 58.48&1517.21 & 815.87& 0.91& 3.34 \\ 
		&& 100&2466.44 &153.20&9836.22 & 992.49& 8.96& 4.84 \\ 
		
		\hline
		\multirow{4}{*}{Quicksort}
				& \multirow{4}{*}{Sorted} 
				& 5& 16.62 & 2.60&16.90 & 2.79& 26.55& 48.17 \\ 
				&& 10& 42.05 & 4.67&43.19 & 5.29& 53.83& 60.25 \\ 
				&& 50& 358.57&25.43&343.96 & 21.16& 176.65& 61.70 \\ 
				&& 100& 843.57 & 63.64&810.67 & 44.09& 245.37& 78.74 \\ 
				\cline{2-9}
		& \multirow{3}{*}{Reversed}  
		& 5& 16.46 & 2.43&16.37 & 3.01& 31.13& 53.62 \\ 
		&& 10& 42.45 & 4.83&43.11 & 5.44& 60.17& 53.58 \\ 
				&& 50& 356.89 & 25.17&350.91 & 23.44& 173.83& 66.61 \\ 
		&& 100& 846.15 & 50.54&826.82 & 42.72& 261.82& 77.31 \\ 
		\cline{2-9}
		& \multirow{3}{*}{Gaussian}    
		& 5& 16.50 & 3.00&17.02 & 3.22& 74.46& 143.95 \\ 
		&& 10& 42.57 & 5.03&42.97 & 4.78& 181.86& 207.35 \\ 
		&& 50& 352.08 & 26.16&349.03 & 22.86& 663.98& 273.06 \\ 
		&& 100& 854.65 & 68.18 &812.52 & 44.76& 883.20& 233.68 \\ 
		\cline{2-9}
		& \multirow{2}{*}{Random}  
		& 5& 16.16 & 2.44&16.87 & 3.13& 30.96& 51.06 \\ 
		&& 10& 43.83 & 5.32 &42.77 & 4.92& 50.13& 52.02 \\ 
		&& 50& 358.22 & 25.25&348.83 & 19.99& 181.51& 60.64 \\ 
		&& 100& 846.99 & 63.46&816.34 & 42.29& 255.82& 74.84 \\ 
		\hline
	\end{tabular}
\end{table*}

\bibliographystyle{abbrv}
\bibliography{literature} 
\end{document}